# Nonlinear Filter Based Image Denoising Using AMF Approach

*\*T.K.Thivakaran*
\*Asst.professor, Department of Information Technology,
Sri Venkateswara College of Engineering,
Post bag.No.3, Pennalur, Sri perumbudhur - 602105. .

*\*\*Dr.RM.Chandrasekaran*
\*\* Professor, Department of CSE,
Anna University,
Trichy.

***Abstract:*** **This paper proposes a new technique based on nonlinear Adaptive Median filter (AMF) for image restoration. Image denoising is a common procedure in digital image processing aiming at the removal of noise, which may corrupt an image during its acquisition or transmission, while retaining its quality. This procedure is traditionally performed in the spatial or frequency domain by filtering. The aim of image enhancement is to reconstruct the true image from the corrupted image. The process of image acquisition frequently leads to degradation and the quality of the digitized image becomes inferior to the original image. Filtering is a technique for enhancing the image. Linear filter is the filtering in which the value of an output pixel is a linear combination of neighborhood values, which can produce blur in the image. Thus a variety of smoothing techniques have been developed that are non linear. Median filter is the one of the most popular non-linear filter. When considering a small neighborhood it is highly efficient but for large window and in case of high noise it gives rise to more blurring to image. The Centre Weighted Median (CWM) filter has got a better average performance over the median filter [8]. However the original pixel corrupted and noise reduction is substantial under high noise condition. Hence this technique has also blurring affect on the image. To illustrate the superiority of the proposed approach by overcoming the existing problem, the proposed new scheme (AMF) Adaptive Median Filter has been simulated along with the standard ones and various performance measures have been compared.**

## INTRODUCTION

In an image, edges and fine details are high frequency content and carry very important information for visual perception. Filters having good edge and image detail preservation properties are highly suitable for digital image filtering. When one would like to remove the noise, it follows certain filtering operation where the signal has to be passed through a filter and the filter in turn removes the undesirable components. Many linear and nonlinear filtering algorithms have been proposed for image denoising [5].

### A. Noise Models:

When it has been discussed on noise it can be getting introduced in the image, either at the time of image generation (or) at the time of image transmission. Noise having Gaussian-like distribution is very often encountered in acquired data.

Generally, Gaussian noise is added to every part of the image and it affects each pixel in the image from its original value by a (usually) small amount based on noise standard deviation. Gaussian noise can easily be removed by locally averaging the pixels inside the window and replace the processing pixel with this average value. Another kind of noise that is present during the image transmission is Salt-Pepper noise [1].It appears as black and/or white impulse of the image. Third type of noise is Impulse noise is classified as fixed valued impulse noise and random valued impulse noise. Generally, impulse noise corrupts certain pixels in the image by either a very low value (Smin) or very high (Smax) intensity value with or without equal probability.

### B. Image Filters

Elimination of noise is one of the major works to be done in computer vision and image processing, as noise leads to the error in the image. Presence of noise is manifested by undesirable information, which is not at all related to the image under study, but in turn disturbs the information present in the image. It is translated into values, which are getting added or subtracted to the true gray level values on a gray level pixel. A digital filter [8] [9] is used to remove noise from the degraded image. As any noise in the image can be result in serious errors. Noise is an unwanted signal, which is manifested by undesirable information. Thus the image, which gets contaminated by the noise, is the degraded image and using different filters can filter this noise. Thus filter is an important subsystem of any signal processing systems. Thus filters are used for image enhancement, as it removes undesirable signal components from the signal of interest. Filters are of different type i.e. linear filters or nonlinear filters. Linear filters generally blur sharp edges, destroy lines and other fine details present in the image. To overcome the problem of linear filtering, non-linear filtering techniques become popular as an alternative to preserve signal structure. Order Statistic filters are one of the most important families of nonlinear image filters that have shown to posses excellent robustness properties in presence of impulsive noise while preserving the edge information. The median filter is the most popular order statistics filter first suggested by Tukey in 1971.It does not posses the drawbacks of linear filters and can effectively eliminate the effects of impulsive noise while preserving the edge information.

PROBLEM FORMULATION

As it had seen that noise elimination is a main concern in computer vision and image processing. Noise presence is manifested by undesirable information, not related to the scene under study, which perturbs the information relative to the form observable in the image. It is translated to more or less severe values, which are added (or) subtracted to the original values on a number of pixels. Noise is of many types. Thus image noise can be Gaussian, Uniform or impulsive distribution. Here we will discuss about the, impulse noise. This impulse noise can be eliminated or the degraded image can be enhanced by the use of advance filter. Due to certain disadvantages of linear filters, nonlinear method of filtering has been proposed in this paper. Nonlinear filter can be very effective in removing the impulse noise. The median filter is the most popular order statistics filter first suggested by Tukey in 1971.It does not posses the drawbacks of linear filters and can effectively eliminate the effects of impulsive noise while preserving the edge information.

It replaces current pixel to be processed by median value of a filtering window around the pixel. Normally, impulse noise has high or low magnitude and is isolated. When we sort pixels in the moving window, noise pixels are usually at the ends of the array. Several techniques have been proposed which try to take the advantage of the average performance of the median filter, either to evaluate noise density, set up parameters or to guide the filtering process. Since the median value must actually be the value of one of the pixels in the neighborhood, the median filter does not create new unrealistic pixel values when the filters overlap an edge. One of the major problems with the median filter is that it is relatively expensive and complex to compute. To find the median it is necessary to sort all the values in the neighborhood into numerical order and this is relatively slow, even with fast sorting algorithms such as quick sort.

The weighted median (WM) filter [3] is an extension of the median filter. The basic idea is to give higher weight to some samples, according to their position with respect to the center of the window. Generally, weights are integers, they specify how many times a sample is replicated in the ordered array . The weighted median filter structure with weights as a=(a1,a2,a3.........ai) and the inputs x=(X1,X2,X3........Xi) is given by Weight Med(X1,X2,X3........Xi) = MED{(a1 ◊X1, a2◊X2, ...ai◊ Xi)} where ◊ is the replication operator.

A special case of WM filter is called center weighted median (CWM) filter [6]. This filter gives more weight only to the central pixel of a window. This leads to improved detail preservation properties at the expense of lower noise suppression. Some of the impulses may not be removed by the filter.

### C. Solution Methodology

A nonlinear filter namely the Adaptive Median Filter (AMF) is proposed to reduce impulsive-like noise while modifying the gray levels of the image as little as possible, resulting in a maximum preservation of the original information [4]. In this paper thrust has been made to devise a filtering scheme to remove impulse noise from images such that the scheme should work at high noise conditions and should perform superior to the existing schemes in terms of noise rejection and retention of original image properties. The detection scheme is devised keeping the CWM filters in mind whereas the median filter is used for the filtering operation for detected noisy pixels. Extensive simulation has been carried out to compare the performance of the proposed filter with other standard schemes. Since impulse noise is not uniformly distributed across the image, it is desirable to replace the Corrupted ones through a suitable filter. For this purpose, a preprocessing is required to detect the corrupted location prior to filtering.

### D. Proposed Algorithms

#### Algorithm : AMF Filter

$S_{i,j}$=filtering window and $Y_{i,j}$=Center pixel in the window

Smin=Minimum gray level value in the filtering window

Smed =Median gray level value in the filtering window

Step1. Initialize W=3

Step2. Compute Smin, Smed and Smax, which are minimum, median and maximum of pixel values in $S_{i,j}$, respectively.

Step3. If Smin < Smed < Smax, then go to step 5. Otherwise, Set W =W+2 until the maximum allowed size is reached.

Step4. If W ⩽Wmax, go to step 2. Otherwise, we replace $Y_{i,j}$ by Smed.

Step5. If Smin < $Y_{i,j}$ < Smax, then $Y_{i,j}$ is not a noise candidate, else we replace $Y_{i,j}$ by Smed. Filter to the test filter in the window W.

Step6. Stop.

### E. Mathematical Analysis

To assess the performance of the proposed filters for removal of impulse noise and to evaluate their comparative Performance, standard performance indices are defined as follows:

**i) Peak Signal to Noise Ratio (PSNR):** It is measured in decibel (dB) and for gray scale image it is defined as:

$$PSNR = 10 \log_{10} \frac{255^2}{mse} \text{ dB}$$

Where mse is the mean square error between the original and the denoised image with size I×J. The higher the PSNR in the restored image, the better is its quality.

**ii) Percentage of Noise Attenuated (PONA):**
It may be defined as the number of pixels getting improved after being filtered.

PONA = (Number of noisy pixels getting improved / Total number of noisy pixels) × 100.

This parameter reflects the capability of the impulse noise detector used prior to filtering.

## RESULTS AND DISCUSSION

The quantitative results has been given in table [Table I & Table II] for the standard ELINA image, for different Percentage of noise, starting from 5% to 30% with a step of 5%, and the comparative analysis has been presented in figure [fig. 1 and fig. 2] for both ELINA and PEPPER image showing the performance of proposed AMF filters over other median filters with 30% of Impulse noise and 60% Pepper Noise respectively. To have a quick insight into the comparative performance of the existing filters along with the proposed one i.e. AME filter, the graphs [Graph1 & Graph2] has also been given for all quantitative measures.

TABLE I. % OF PSNR VALUE FOR ELAINE IMAGE

| % Impulse Noise | CWMF | AMF |
|---|---|---|
| 5 | 98.7562 | 99.757 |
| 10 | 98.8724 | 99.8234 |
| 15 | 97.572 | 99.7087 |
| 20 | 97.90 | 99.5008 |
| 25 | 96.1055 | 99.2512 |
| 30 | 95.5977 | 99.2028 |

TABLE II. % OF NOISE ATTENUATED IN ELIAINE IMAGE

| % Impulse Noise | CWMF | AMF |
|---|---|---|
| 5 | 27.4687 | 28.4385 |
| 10 | 26.5805 | 27.7254 |
| 15 | 24.9343 | 27.312 |
| 20 | 24.0316 | 26.2449 |
| 25 | 22.2725 | 24.9492 |
| 30 | 21.3819 | 24.3714 |

GRAPH I. % OF PSNR VALUE FOR ELIAINE IMAGE

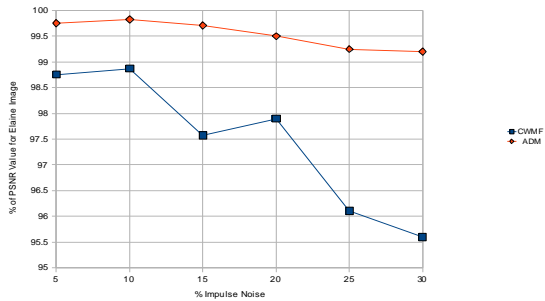

GRAPH II. % OF NOISE ATTENUATED IN ELIAINE IMAGE

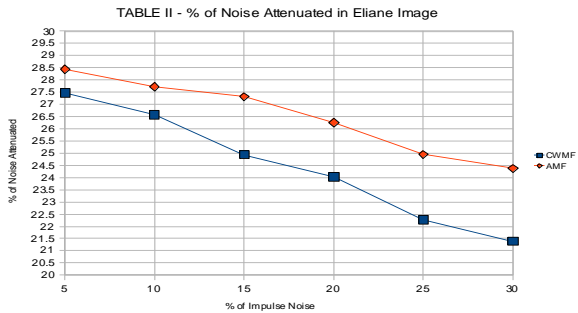

FIG I. COMPARISONS OF DIFFERENT MEDIAN BASED FILTERS USED FOR ELAINE IMAGE

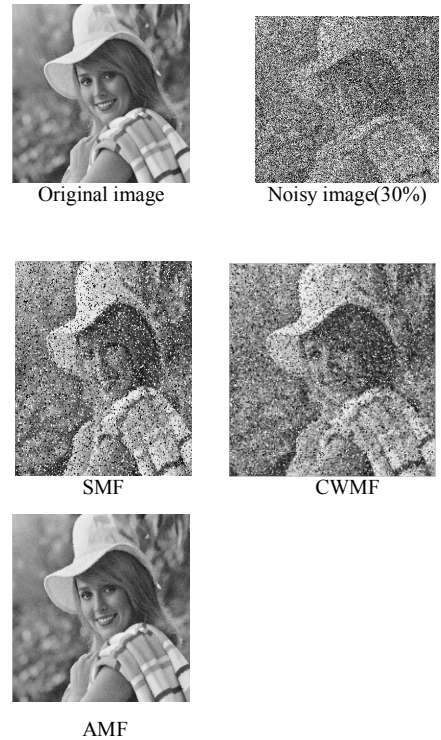

FIG II. COMPARISONS OF DIFFERENT MEDIAN BASED FILTERS USED FOR PEPPER IMAGE

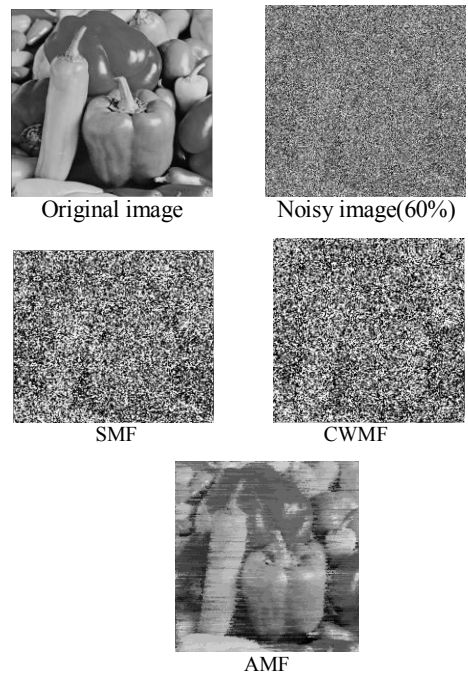

### ACKNOWLEDGMENT

The authors wish to thank Mr.Francis & Mr.Durai, Department of I.T, SVCE for their kind support.

### CONCLUSION AND FUTURE ENHANCEMENT

This paper proposed new non-linear filters to remove the impulse noise from the images. To illustrate the efficiency of the proposed AMF schemes, we have simulated the new schemes along with the existing ones and various restored measures have been compared. All the filtering techniques have been simulated in MATLAB 7.1 with Pentium-IV processor. The schemes are simulated using standard images ELIANA and PEPPER. The impure contaminations include Salt and Pepper noise. The proposed schemes AMF filter is found to be superior, i.e. better results than other parameter for restoration compared to the existing schemes.

It is expected that future research will focus on developing EHW architecture to filter the noise in the images.

### VI.  BIOGRAPHICS

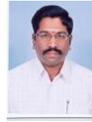

1. Mr. T.K.Thivakaran is presently   a research scholar in MS university, Thirunelveli in the faculty of Computer Science and Engineering. He is working as Assistant Professor in the faculty of Information Technology, Sri Venkateswara college of Engineering, Chennai. His area of research includes Image Processing, Cryptography and Network Security.

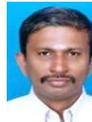

2. Dr. RM. Chandrasekaran is presently working as Registrar, Anna University, Trichy. He has published more than 20 papers in National and International journals.  His research areas of interest include Image Processing, Mobile Ad-hoc network and Network Security and Wireless networks.